\documentclass{article}
\usepackage{ijcai18}

\usepackage{times}
\usepackage{xcolor}
\usepackage{soul}
\usepackage[utf8]{inputenc}
\usepackage[small]{caption}
\usepackage{epsfig}
\usepackage{graphicx}
\usepackage{amsmath}
\usepackage{amssymb}
\usepackage{algorithmic} 
\usepackage{mathrsfs}
\usepackage{subfigure}
\usepackage{paralist}

\usepackage[linesnumbered,ruled]{algorithm2e}

\title{Exploiting Images for Video Recognition with Hierarchical Generative Adversarial Networks  }

\author{
	Feiwu Yu$^1$, 
	Xinxiao Wu$^1$ \thanks{Corresponding author: Xinxiao Wu}, 
	Yuchao Sun$^1$, 
	Lixin Duan$^2$ 
	\\ 
	$^1$ Beijing Laboratory of Intelligent Information Technology, School of Computer Science,\\Beijing Institute of Technology \\
	$^2$ Big Data Research Center, University of Electronic Science and Technology of China\\
	$\{$yufeiwu,wuxinxiao,sunyuchao$\}$@bit.edu.cn,
	lxduan@uestc.edu.cn
}

\begin{document}

\maketitle

\begin{abstract}
Existing deep learning methods of video recognition usually require a large number of labeled videos for training. 
But for a new task, videos are often unlabeled and it is also time-consuming and labor-intensive to annotate them.
Instead of human annotation, we try to make use of existing fully labeled images to help recognize those videos. 
However, due to the problem of domain shifts and heterogeneous feature representations, the performance of classifiers trained on images may be dramatically degraded for video recognition tasks. In this paper, we propose a novel method, called Hierarchical Generative Adversarial Networks (HiGAN), to enhance recognition in videos (i.e., target domain) by transferring knowledge from images (i.e., source domain). The HiGAN model consists of a \emph{low-level} conditional GAN and a \emph{high-level} conditional GAN. By taking advantage of these two-level adversarial learning, our method is capable of learning a domain-invariant feature representation of source images and target videos. Comprehensive experiments on two challenging video recognition datasets (i.e. UCF101 and HMDB51) demonstrate the effectiveness of the proposed method when compared with the existing state-of-the-art domain adaptation methods.
\end{abstract}

\section{Introduction}
\begin{figure*}
	\centering
	\includegraphics[width=0.95\linewidth]{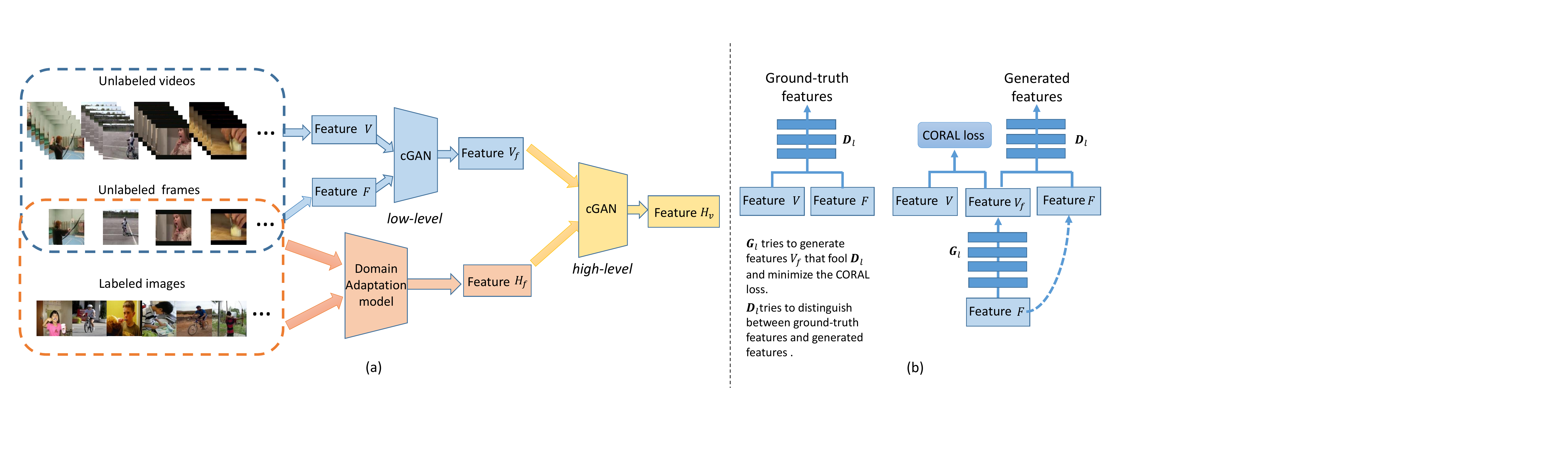}
	\caption{(a) The framework of the proposed HiGAN. We first train a Domain Adaptation model to learn features \emph{$H_f$}. Then, we learn the \emph{low-level} condition GAN with features \emph{V} and \emph{F}, and calculate the features \emph{$V_f$}. After that, we train the \emph{high-level} condition GAN with features \emph{$V_f$} and \emph{$H_f$}, and calculate the features \emph{$H_v$}.   (b) The training process of the \emph{low-level} conditional GAN (the \emph{high-level} conditional GAN has a similar structure).}
	\label{fig:framework}
\end{figure*}

Video recognition is an active research area because of its wide applications such as anomalous events detection, action retrieval, human behavior analysis and so forth. Thanks to the success of deep neural networks, the performance of video recognition has been dramatically improved. However, training deep video classifiers requires collecting and labeling a large number of videos, which is often time-consuming and labor-intensive. On the other hand, images are much easier and cheaper to collect and annotate, and there are also many existing labeled image datasets which can be utilized. Furthermore, the computational cost of learning deep classifiers of images is much less than that of videos, so it would be beneficial a lot to transfer knowledge from source images to target videos \cite{DSM,zhang2017semi,gan2016webly}.

However, applying the images trained classifier to video data directly introduces the
domain shift problem, where the variations in data between source and target significantly degrade the recognition performance at test time. To solve this problem, many domain adaptation methods have been well explored. 
A rich line of prior work has focused on learning shallow features by obtaining a symmetric transformation and minimizing a distance metric of domain discrepancy \cite{KCCA,DAMA,DSM,HEMAP,HFA,CDLS,zhang2017semi}. Recently, studies have shown that deep neural networks can learn more transferable features for domain adaptation  \cite{TNT,DAN,RTN,JAN}.  In order to represent videos more appropriately, spatiotemporal features that are totally different from image representations with different feature dimensions and physical meanings are usually extracted, which makes the problem worse.

To tackle these issues aforementioned, inspired by recent advances in generative adversarial networks (GANs) \cite{gan}, we propose a new approach referred to as \textit{Hierarchical Generative Adversarial Networks} (HiGAN) to transfer knowledge from images to videos by learning domain-invariant feature representations between them. As illustrated in Figure \ref{fig:framework}(a), our approach mainly consists of two components: a two-level hierarchical conditional GAN model and a domain adaptation model. The domain adaptation model is adopted to learn a common feature representation between source images and target video frames, called image-frame feature. The two-level HiGAN is designed to have a \emph{low-level} conditional GAN and a \emph{high-level} conditional GAN. The \emph{low-level} conditional GAN is built to connect videos and their corresponding video frames by learning a mapping function from frame features to video features in the target domain. The \emph{high-level} conditional GAN, on the other hand, is modeled to bridge the gap between source images and target videos by formulating a mapping function from video features to image-frame features. Therefore, any classifiers trained on the source image domain can be effectively transferred to the target video domain for recognition, with the help of the transformable features between them. In HiGAN, the Correlation Alignment (CORAL) Loss \cite{dcoral} is utilized to minimize the difference in second-order statistics between the generated features and the real features.

It is worth emphasizing that, by exploiting the correspondence between a video and its related frames, the projected video features, which keep the temporal motion of videos, can be learned with the absence of any paired image-video training examples in an unsupervised scenario.
The experiments conducted on the UCF101 and HMDB51 datasets demonstrate that our method can achieve better results than the current state-of-the-art domain adaptation methods.

\section{Related Work}
To harness the information from large-scale image datasets, several
works use images as auxiliary training data for video recognition \cite{DSM,zhang2017semi,gan2016webly}. In \cite{DSM}, they proposed a new multiple domain adaptation method for event recognition  by leveraging a large number of loosely labeled web images from different sources. In \cite{gan2016webly}, they jointly exploited source images and source videos for labeling-free video recognition, and proposed a mutually voting approach to filter noisy source images and video frames. In \cite{zhang2017semi}, they presented a classifier of image-to-video adaptation, which borrows the
knowledge adapted from images, and utilizes the heterogeneous features of unlabeled videos to enhance the performance of action recognition. Different from these methods, our method takes the temporal information of videos into consideration without extracting keyframes from videos.

When transferring knowledge between different domains, the domain shift will cause the classifier learned from the source domain to perform poorly on the target domain. Numerous domain adaptation methods have been well explored where the key focus is to learn domain-invariant feature representations. A common strategy is to find a mapping function that would align the source distribution with the target domain \cite{long2014transfer,long2015domain,sun2016return,LRSR,Busto_2017_ICCV,KCCA,DAMA,HEMAP,HFA,CDLS}. Recently, deep neural networks have been exploited for domain adaptation \cite{DAN,RTN,TNT,JAN,Carlucci_2017_ICCV}, In \cite{DAN,RTN,JAN}, Long \textit{et al.} embedded domain-adaptation modules into deep networks in which all the layers corresponding to task-specific features are adapted in a layerwise manner. These methods focus on the image-to-image adaptation.
In \cite{TNT}, the architecture of Transfer Neural Trees has been presented, which jointly solves the cross-domain feature adaptation and classification between heterogeneous domains in a semi-supervised scenario. In contrast, our method can transfer knowledge between images and videos in an unsupervised scenario.

Generative Adversarial Networks (GANs) \cite{gan} are a class of artificial intelligence algorithms that are implemented by a system of two competing models: a generative model and a discriminative model. These two networks compete with each other in a two-player minimax game: the generator is learning to produce as realistic as possible samples at confusing the discriminator, and the discriminator is learning to get as correct as possible results at distinguishing generated samples from real data. GANs have achieved impressive progress in domain adaptation \cite{Bousmalis_2017_CVPR,sankaranarayanan2017generate}. In \cite{sankaranarayanan2017generate}, they proposed an approach that brings the source and target distributions closer in a learned joint feature space. In \cite{Bousmalis_2017_CVPR}, they presented
a new approach that learns a transformation in the pixel space from one domain to another. Different from these homogeneous domain adaptation methods based on single GAN, our hierarchical GAN method can effectively learn a common feature representation shared by heterogeneous domains.

\section{Hierarchical Generative Adversarial Networks}

\subsection{Model}
Given a source domain with a set of labeled images and a target domain with a set of unlabeled videos that share the same categories, our goal is to learn a common feature representation shared by source and target domains, on which the classifier built from the source domain adapts well to the target domain. Since a video is composed of a sequence of frames, we assume that a short video clip has a relationship with any frames in it, which provides a natural correlation between frame features and video features. On the other hand, video frames are a collection of images, which could be easily adapted to the source image domain. We explore the image-to-video adaptation by leveraging the relationship among source images, target video clips and their corresponding frames.

\paragraph{Image-Frame Feature.}Let $\mathcal{D}_s={\{\mathbf{x}_s^i,y_s^i\}}_{i=1}^{n_s}$ represent a labeled dataset of $n_s$ images from the source domain and $\mathcal{D}_t=\{\mathbf{x}_t^j\}_{j=1}^{n_t}$ denote an unlabeled dataset of $n_t$ videos from the target domain. We divide each video into several clips with the same length, building up an unlabeled video clip domain $\mathcal{D}_{v}=\{\mathbf{x}_{v}^k\}_{k=1}^{n_v}$, where $n_v$ is the number of video clips in total. For each clip, a frame is randomly selected and all the selected frames compose an unlabeled video frame domain $\mathcal{D}_{f}=\{\mathbf{x}_{f}^k\}_{k=1}^{n_v}$. Since both of $\mathcal{D}_s$ and $\mathcal{D}_{f}$ are image collections, we adopt a state-of-the-art deep domain adaptation model (i.e. JAN \cite{JAN}) to learn the common feature (called image-frame feature) of source images and target frames. In the common image-frame feature space, the source images and the target frames are represented by $\boldsymbol{H}_{s}=[\mathbf{h}_{s}^1, \mathbf{h}_{s}^2,\cdots,\mathbf{h}_{s}^{n_s}]\in \mathbb{R}^{n_s\times d_h}$ and $\boldsymbol{H}_{f}=[\mathbf{h}_{f}^1, \mathbf{h}_{f}^2,\cdots,\mathbf{h}_{f}^{n_v}]\in \mathbb{R}^{n_v\times d_h}$, respectively, where $d_h$ is the feature dimension.

\paragraph{Frame-to-Video Mapping.}For the $\mathcal{D}_{v}$, we extract the C3D feature \cite{tran2015learning} to describe the target video clips as $\boldsymbol{V}=[\mathbf{v}^1, \mathbf{v}^2,\cdots,\mathbf{v}^{n_v}]\in \mathbb{R}^{n_v\times d_v}$, where $d_v$ is the feature dimension. For the $\mathcal{D}_{f}$, we employ the ResNet \cite{he2016deep} to extract deep features of target frames as
$\boldsymbol{F}=[\mathbf{f}^1, \mathbf{f}^2,\cdots,\mathbf{f}^{n_v}]\in \mathbb{R}^{n_v\times d_f}$, where $d_f$ is the feature dimension. Generally speaking, frame features and video features are from two heterogeneous feature spaces, which means $d_v\ne d_f$. Considering the correspondence between a video frame and its related video clip, we get a collection of $n_v$ instances of video-frame pairs, denoted as $P=\{p_k\}_{k=1}^{n_v}$ with $p_k=(\mathbf{v}^k, \mathbf{f}^k)$. We aim at learning a mapping function from video frames to their related video clips. Specifically, the frame features $\boldsymbol{F}$ are projected to video clip features $\boldsymbol{V}$ as $\boldsymbol{V}_{f}=G_{l}(\boldsymbol{F};\theta_{G_{l}})$, where $G_{l}(\cdot;\theta_{G_{l}})$ is the mapping function and $\boldsymbol{V}_{f}=[\mathbf{v}_{f}^1, \mathbf{v}_{f}^2,\cdots,\mathbf{v}_{f}^{n_v}]\in \mathbb{R}^{n_v\times d_v}$ indicates the generated video feature from $\boldsymbol{F}$.

\paragraph{Video-to-Image Mapping.} For each frame $\mathbf{x}_f^k$, it can be represented by two different features  $\mathbf{v}_f^k\in \boldsymbol{V}_f$ and $\mathbf{h}_f^k\in \boldsymbol{H}_f$, thus we have a collection of $n_v$ two heterogeneous features pairs, expressed as $Q=\{q_k\}_{k=1}^{n_v}$ with $q_k=(\mathbf{h}_{f}^k,\mathbf{v}_{f}^k)$. Given $Q$, a projection $\boldsymbol{H}^{'}_{f}=G_{h}(\boldsymbol{V}_{f};\theta_{G_{h}})$ from $\boldsymbol{V}_f$ to $\boldsymbol{H}_f$ can be learned. Here, $G_{h}(\cdot;\theta_{G_{h}})$ is the mapping function and $\boldsymbol{H}^{'}_{f}$ is the generated image-frame feature from $\boldsymbol{V}_f$.
Note that $\boldsymbol{H}_f$ and $\boldsymbol{H}_s$ share the same feature space learned by JAN model. In fact, $\boldsymbol{V}_f$ and $\boldsymbol{V}$ also come from the same video feature space. Consequently, $G_{h}(\cdot;\theta_{G_{h}})$ can be actually considered as the projection from videos to images, and the video clip features
$\boldsymbol{V}$ can be projected to the image-frame space as $\boldsymbol{H}_{v}=G_{h}(\boldsymbol{V};\theta_{G_{h}})$, where $\boldsymbol{H}_{v}=[\mathbf{h}_{v}^1,\mathbf{h}_{v}^2,\cdots,\mathbf{h}_{v}^k]\in \mathbb{R}^{n_v\times d_h}$ indicates the generated image-frame feature from $\boldsymbol{V}$. Followed by averaging the clip features of each video, we obtain the final features for videos, denoted as $\boldsymbol{H}_{t}=[\mathbf{h}_{t}^1, \mathbf{h}_{t}^2,\cdots,\mathbf{h}_{t}^{n_t}]\in \mathbb{R}^{n_t\times d_h}$, which can be directly compared against source image features $\boldsymbol{H}_s$.

Motivated by the success of GANs in multiple fields, we propose a HiGAN model where the \emph{low-level} GAN and the \emph{high-level} GAN are designed to learn the mapping functions $G_l$ and $G_h$, respectively. Our whole objective includes two terms. One is the adversarial loss \cite{gan} for matching the distribution of generated features to the data distribution in the original domain. The other is the CORAL loss \cite{dcoral} for minimizing the difference in second-order statistics between the synthesized features and the original features.

\subsection{Loss}
\paragraph{Adversarial Loss.}For the mapping function $G_{l}$ associated with the discriminator $D_{l}$, we denote the data distribution as $\boldsymbol{V}\sim P_{data}(\boldsymbol{V})$ and $\boldsymbol{F}\sim P_{data}(\boldsymbol{F})$, then the objective is formulated as:
\begin{equation}
\begin{split}
\mathcal{L}_{GAN}&(D_{l},G_{l},\boldsymbol{F},\boldsymbol{V})=
E_{\boldsymbol{V}\sim P_{data}(\boldsymbol{V})}[\log D_{l}(\boldsymbol{V}|\boldsymbol{F})]\\
&+E_{\boldsymbol{F}\sim P_{data}(\boldsymbol{F})}[\log (1-D(G_{l}(\boldsymbol{F})|\boldsymbol{F}))] ,
\label{eqn:cGANloss1}
\end{split}
\end{equation}
where $G_{l}$ attempts to generate video features $G_{l}(\boldsymbol{F})$ that resemble the video features from $\boldsymbol{V}$, while $D_{l}$ tries to distinguish between generated features $G_{l}(\boldsymbol{F})$, and ground-truth features $\boldsymbol{V}$. In other words, $G_{l}$ aims at minimizing this objective against an adversary $D_{l}$ that tries to maximize it, formulated by  $\min_{G_{l}}\max_{D_{l}}\mathcal{L}_{GAN}(D_{l},G_{l},\boldsymbol{F},\boldsymbol{V})$. We introduce a similar adversarial loss for $G_{h}$ and $D_{h}$ , given by  $\min_{G_{h}}\max_{D_{h}}\mathcal{L}_{GAN}(D_{h},G_{h},\boldsymbol{V}_{f},\boldsymbol{H}_{f})$.

\paragraph{CORAL Loss.} It has been proven that the GAN objective combined with another loss could produce significant good results \cite{Zhu_2017_ICCV}. In our method, we introduce the CORAL loss \cite{dcoral} to minimize the difference in second-order statistics between the generated features and the real features. The CORAL Loss is simple, effective and can be easily integrated into a deep learning architecture.

For the generator $G_{l}$ associated with $D_{l}$, $\boldsymbol{V}$ is the real features and $\boldsymbol{V}_{f}$ is the synthesized features. Suppose $\boldsymbol{E}_{v}$ and $\boldsymbol{E}_{v_{f}}$ denote the feature covariance matrices. The CORAL loss as the distance between the second-order statistics (covariance) of $\boldsymbol{V}$ and $\boldsymbol{V}_{f}$ is expressed as follows:
\begin{equation}
\begin{split}
\mathcal{L}_{CORAL}(\boldsymbol{V},\boldsymbol{V}_{f})=\frac{1}{4d_v^2}\|\boldsymbol{E}_{V}-\boldsymbol{E}_{V_{f}}\|_F^2 ,
\label{eqn:CORALloss}
\end{split}
\end{equation}
where $\|\cdot\|_F^2$ denotes the squared matrix Frobenius norm. The covariance matrices $\boldsymbol{E}_{V}$ and $\boldsymbol{E}_{V_f}$are calculated by

\begin{equation}
\begin{split}
\boldsymbol{E}_{V}=\frac{1}{n_v-1}(\boldsymbol{V}^\mathrm{T}\boldsymbol{V}-\frac{1}{n_v}(\boldsymbol{1}^\mathrm{T}\boldsymbol{V})^\mathrm{T}(\boldsymbol{1}^\mathrm{T}\boldsymbol{V})) ,
\label{eqn:covariance1}
\end{split}
\end{equation}

\begin{equation}
\begin{split}
\boldsymbol{E}_{V_{f}}=\frac{1}{n_v-1}(\boldsymbol{V}_{f}^\mathrm{T}\boldsymbol{V}_{f}-\frac{1}{n_v}(\boldsymbol{1}^\mathrm{T}\boldsymbol{V}_{f})^\mathrm{T}(\boldsymbol{1}^\mathrm{T}\boldsymbol{V}_{f})) ,
\label{eqn:covariance2}
\end{split}
\end{equation}
where $\boldsymbol{1}$ is a column vector with all elements equal to 1.

For the generator $G_{h}$ associated with $D_{h}$, we introduce a similar CORAL loss as well:
\begin{equation}
\begin{split}
\mathcal{L}_{CORAL}(\boldsymbol{H}_{f},\boldsymbol{H}^{'}_{f})=\frac{1}{4d_h^2}\|\boldsymbol{E}_{{H}_{f}}-\boldsymbol{E}_{{H}^{'}_{f}}\|_F^2 .
\label{eqn:CORALloss1}
\end{split}
\end{equation}
\subsection{Objective}
In order to prevent the learned parameters from overfitting, we introduce the regularization term:
\begin{equation}
\begin{split}
\mathcal{L}_{reg}(D_{l},G_{l})=\sum_{l_D=1}^{L_D}\|W_{D_{l}}^{l_D}\|_F+\sum_{l_G=1}^{L_G}\|W_{G_{l}}^{l_G}\|_F ,
\label{eqn:regLoss1}
\end{split}
\end{equation}

\begin{equation}
\begin{split}
\mathcal{L}_{reg}(D_{h},G_{h})=\sum_{l_D=1}^{L_D}\|W_{D_{h}}^{l_D}\|_F+\sum_{l_G=1}^{L_G}\|W_{G_{h}}^{l_G}\|_F,
\label{eqn:regLoss2}
\end{split}
\end{equation}
where $W_{D_{l}}^{l_D}$, $W_{G_{l}}^{l_G}$, $W_{D_{h}}^{l_D}$ and $W_{G_{h}}^{l_G}$ represent the layer-wise parameters of networks. $L_D$ and $L_G$ denote layer numbers of discriminator and generator, respectively.
Based above, the whole objective becomes
\begin{equation}
\begin{split}
\mathcal{L}(D_{l},G_{l})&=\lambda_1\mathcal{L}_{cGAN}(D_{l},G_{l},\boldsymbol{F},\boldsymbol{V})\\
&+\lambda_2 \mathcal{L}_{CORAL}(\boldsymbol{V},\boldsymbol{V}_{f})\\
&+\mathcal{L}_{reg}(D_{l},G_{l}),
\label{eqn:fullLoss1}
\end{split}
\end{equation}

\begin{equation}
\begin{split}
\mathcal{L}(D_{h},G_{h})&=\lambda_3\mathcal{L}_{cGAN}(D_{h},G_{h},\boldsymbol{V}_{f},\boldsymbol{H}_{f})\\
&+\lambda_4 \mathcal{L}_{CORAL}(\boldsymbol{H}_{f},\boldsymbol{H}^{'}_{f})\\
&+\mathcal{L}_{reg}(D_{h},G_{h}),
\label{eqn:fullLoss2}
\end{split}
\end{equation}
where $\lambda_1$, $\lambda_2$, $\lambda_3$ and $\lambda_4$ are weight parameters that control the relative importance of adversarial loss and CORAL loss, respectively.

\subsection{Algorithm}
To learn the optimal feature representations, the generator $G$ and the discriminator $D$ compete with each other in a two-player minimax game, jointly minimizing the adversarial loss in Eq. (\ref{eqn:cGANloss1}) and the CORAL loss in Eq. (\ref{eqn:CORALloss}). We aim to solve:
\begin{equation}
\begin{split}
D_{l}^*,G_{l}^*=arg\min_{G_{l}}\max_{D_{l}}\mathcal{L}(D_{l},G_{l}) ,
\label{eqn:opt1}
\end{split}
\end{equation}

\begin{equation}
\begin{split}
D_{h}^*,G_{h}^*=arg\min_{G_{h}}\max_{D_{h}}\mathcal{L}(D_{h},G_{h}) .
\label{eqn:opt2}
\end{split}
\end{equation}
The algorithm of HiGAN  is summarized in Algorithm 1.
\begin{algorithm}
	\small
	\label{alg:alg1}
	\caption{Hierarchical Generative Adversarial Networks}
	\KwIn{The source image domain $\mathcal{D}_s$, the target video domain $\mathcal{D}_t$, the video clip domain $\mathcal{D}_v$, the video frame domain $\mathcal{D}_f$, the deep features $\boldsymbol{V}$ of $\mathcal{D}_{v}$, the deep features $\boldsymbol{F}$ of $\mathcal{D}_{f}$ }
	\KwOut{The deep features $\boldsymbol{H}_s$ of $\mathcal{D}_s$ and $\boldsymbol{H}_t$ of $\mathcal{D}_t$}
	Train the JAN model with $\mathcal{D}_s$ and $\mathcal{D}_{f}$ to learn features $\boldsymbol{H}_s$ and $\boldsymbol{H}_{f}$ of $\mathcal{D}_s$ and $\mathcal{D}_f$, respectively.\\
	Learn the mapping function $G_{l}(\cdot;\theta_{G_{l}})$ with deep features $\boldsymbol{F}$ and $\boldsymbol{V}$ via Eq. (\ref{eqn:fullLoss1}), and calculate the new deep feature $V_f$ of $\mathcal{D}_{f}$.\\
	Learn the mapping function $G_{h}(\cdot;\theta_{G_{h}})$ with deep features $\boldsymbol{V}_{f}$ and $\boldsymbol{H}_{f}$ by Eq. (\ref{eqn:fullLoss2}), and calculate the new deep feature $\boldsymbol{H}_{v}$ of $\mathcal{D}_{v}$. \\
	
	Average the clip features $\boldsymbol{H}_{v}$ of each video to obtain the features $\boldsymbol{H}_t$ of $\mathcal{D}_t$.\\
	
	Return $\boldsymbol{H}_s$ and $\boldsymbol{H}_t.$
	
\end{algorithm}

\section{Experiments}

\subsection{Datasets}

To evaluate the performance of our method, we conduct the experiments on two complex video datasets, i.e., UCF101 \cite{soomro2012ucf101} and HMDB51 \cite{stanford40}. For the UCF101 as the target video domain, the source images come from the Stanford40 dataset \cite{stanford40}. For the HMDB51 as the target video domain, the source image domain consists of Standford40 dataset and HII dataset \cite{tanisik2016facial}, denoted by EADs dataset.

\emph{Stanford40 and UCF101} (S$\rightarrow$U): The UCF101 is a dataset of action videos collected from YouTube with 101 action categories. The Stanford40 dataset include diverse action images with 40 action categories. We choose 12 common categories between these two datasets.

\emph{EADs and HMDB51} (E$\rightarrow$H): The HMDB51 dataset has 51 classes, containing 6766 video sequences. The EADs dataset consists of Stanford40 and HII datasets. The HII dataset has 10 action images and each class contains at least 150 images, forming a total of 1972 images. The 13 shared categories between the EDAs and the HMDB51 datasets are adopted in our experiment.

\subsection{Experiment Setup}
We split each target video into 16-frame clips without overlap, and all the clips from all the target videos construct the video-clip domain. The deep feature of each video clip is the 512D feature vector from the pool5 layer of 3D CoveNets \cite{tran2015learning} which are trained on a large-scale video dataset. For each video clip, we randomly sample one frame and all the frames from all the video clips compose the video-frame domain. The deep feature of frames is the 2048D vector extracted from the \emph{pool5} layer of ResNet \cite{he2016deep}.
We utilize the JAN \cite{JAN} method based on the ResNet to get the domain-invariant image-frame features between source images and target video frames from the \emph{pool5} layer of JAN with the dimension of 2048D.

\paragraph{Implementation details.}
To model the two generators in HiGAN, we deploy four-layered feed-forward neural networks activated by \emph{relu} function, (i.e., $2048 \rightarrow 1024\rightarrow 1024\rightarrow 1024\rightarrow 512$ for $G_{l}(f;\theta_{G_{l}})$ and $512 \rightarrow 1024\rightarrow 1024\rightarrow 2048\rightarrow 2048$ for $G_{h}(v_{f};\theta_{G_{h}})$). In terms of two discriminators, we both utilize three fully connected layers ( $2560 \rightarrow 1280\rightarrow 640\rightarrow 1$) activated by \emph{relu} function, except for the last layer.

In practice, we replace the negative log likelihood objective by a least square loss \cite{mao2016multi}, which performs more stably during training and generates higher quality results.

Since the adversarial losses and the CORAL losses have different orders of magnitude according to the experiments, and the \emph{low-level} conditional GAN and the \emph{high-level} one have very similar network structures, we set $\lambda_2=\lambda_4=100$, $\lambda_1=\lambda_3=1$ in Eq. (\ref{eqn:fullLoss1}) and Eq. (\ref{eqn:fullLoss2}) for all the experiments, to somehow balance the two types of losses. We employ the Adam solver \cite{kingma2014adam} with a batch size of 64. All the networks were trained from scratch with the learning rate of 0.00002 for the \emph{low-level} conditional GAN and 0.000008 for \emph{high-level} conditional GAN.

\paragraph{Related methods.}We evaluate the effectiveness of our approach by investigating whether our generated features of source images and target videos are more transferable than other domain adaptation methods, that is, whether the recognition accuracy on the target videos could be improved. Specifically, the compared methods are: 1) MKL \cite{MKL}; TJM \cite{long2014transfer}; TKL \cite{long2015domain}; CORAL \cite{sun2016return}; LRSR \cite{LRSR}; ATI \cite{Busto_2017_ICCV}; KCCA \cite{KCCA}; HEMAP \cite{HEMAP}; DAMA \cite{DAMA}; HFA \cite{HFA}; CDLS \cite{CDLS}, which are traditional shallow domain adaptation methods, and 2)  ResNet \cite{he2016deep}; DAN \cite{DAN}; RTN \cite{RTN}; JAN \cite{JAN}; DAL \cite{Carlucci_2017_ICCV}, which are deep domain adaptation methods. The MKL is considered as a traditional baseline method which directly uses source classifiers in the target domain. Similarly, the ResNet is taken as a deep baseline method. As fair comparison with identical evaluation setting, the image-frame (i.e., JAN) features of source images are taken as source features in the traditional shallow domain adaptation methods. For all the methods, we use the accuracy for performance evaluation.

Note that MKL, TJM, TKL, CORAL, LRSR and ATI can only handle the homogeneous domain adaptation problem, when the data from the source and target domains are with the same type of feature. Therefore, for these methods, each target video is represented by the mean of the image-frame features of all the frames.

For the traditional heterogeneous domain adaptation methods of KCCA, HEMAP, DAMA, HFA and CDLS, the C3D feature is adopted to describe the target videos. Considering that all these methods require the labeled training data in the target domain, we randomly choose three videos per class as the labeled target data and take the rest videos as the test data. For each dataset, we repeat the sampling for 5 times and report the average results.

Regarding the deep methods of ResNet, DAN, RTN, JAN, DAL, since they only take images as input, the source images and target video frames are utilized to train the networks. The output scores (from the last \emph{fc} layer) of all the frames are further averaged to determine the class label of the video. All the deep methods are implemented based on the Caffe framework \cite{jia2014caffe}
and fine-tuned from Caffe-provided models of ResNet, which are pre-trained on the ImageNet 2012 dataset. We fine-tune
all convolutional and pooling layers and train the classifier
layer via back propagation.

\begin{table}
	
	\setlength{\tabcolsep}{3mm}{
		\begin{tabular}{|c|c|c|}
			\hline
			Method & S$\rightarrow$U &E$\rightarrow$H  \\
			\hline
			MKL \cite{MKL} & 0.889& 0.392  \\
			TJM \cite{long2014transfer} & 0.898 & 0.314 \\
			TKL \cite{long2015domain} & 0.902 & 0.391 \\
			CORAL \cite{sun2016return} & 0.904& 0.398  \\
			LRSR \cite{LRSR} & 0.893& 0.380  \\
			ATI \cite{Busto_2017_ICCV} &0.905& 0.330  \\
			\hline
			HiGAN (no labeled video) & \textbf{0.954}& \textbf{0.446}\\
			
			\hline
			KCCA \cite{KCCA} & 0.853 & 0.366 \\
			HEMAP \cite{HEMAP} &0.432 &0.213 \\
			DAMA \cite{DAMA} & 0.860 & 0.438 \\
			HFA \cite{HFA} & 0.885& 0.459  \\
			CDLS \cite{CDLS} & 0.885& 0.435  \\
			\hline
			HiGAN (three labeled videos) & \textbf{0.967}& \textbf{0.526}\\
			
			\hline
	\end{tabular}}
	\caption{Comparison between our method and the traditional shallow domain adaptation methods.}
	\label{tab:table1}       
\end{table}
\begin{table}
	
	\setlength{\tabcolsep}{5.3mm}{
		\begin{tabular}{|c|c|c|}
			\hline
			Method & S$\rightarrow$U &E$\rightarrow$H  \\
			\hline
			ResNet \cite{he2016deep} & 0.814& 0.385  \\
			DAN \cite{DAN} & 0.842 & 0.395 \\
			RTN \cite{RTN} & 0.838 & 0.402 \\
			JAN \cite{JAN} & 0.914& 0.409  \\
			DAL \cite{Carlucci_2017_ICCV} & 0.826& 0.426  \\
			\hline
			HiGAN  & \textbf{0.954}& \textbf{0.446}\\
			\hline
	\end{tabular}}
	\caption{Comparison between our method and the deep domain adaptation methods.}
	\label{tab:table2}       
\end{table}
\subsection{Results}
\paragraph{Comparison with existing domain adaptation methods.}
Table \ref{tab:table1} and Table \ref{tab:table2} show the recognition accuracies of traditional shallow domain adaptation methods and deep domain adaptation methods, respectively. From them, we can notice that our method achieves the best performance compared with other state-of-the-art methods for both datasets, which explicitly demonstrates the effectiveness of transferring knowledge from images to videos for video recognition. The more detailed observations are as follows.

(\textbf{1}) When videos are represented by C3D features which are different from source image features, the traditional heterogeneous methods substantially perform worse than the traditional homogeneous methods on the S$\rightarrow$U dataset, even though they have some labeled videos in the target domain. This shows that it is a quite challenging task to conduct domain adaptation between images and videos when they are respectively represented by different features. Compared with them, our proposed HiGAN can achieve better performance owing to the strengths of taking video frames as a bridge to obtain the common feature shared by heterogeneous images and videos.

(\textbf{2}) When there are no labeled data in the target domain, both traditional methods and deep methods are better than the baseline methods (i.e., MKL and ResNet) on the S$\rightarrow$U dataset, which indicates that these adaptation methods can improve action recognition performance, but the improvement is not significant. On the E$\rightarrow$H dataset, traditional methods slightly underperform the baseline MKL. A possible explanation is that the huge difference between source images and target video frames on the E$\rightarrow$H dataset leads to the negative transfer. On the other hand, deep transfer learning methods are better than the baseline ResNet, which verifies that deep networks are better at addressing negative transfer issue. Compared with them, our proposed HiGAN gains performance improvement on both datasets, which demonstrates that our method can attain respectable performances even though there is a huge difference between source images and target videos.

\paragraph{Analysis of the loss function.}

In Table \ref{tab:table4}, we assess the individual effect of the advesarial loss and the CORAL loss in HiGAN, with
respect to the recognition accuracies in two datasets. We develop two variants of our HiGAN: one with the adversarial loss but without the CORAL loss, the other with the CORAL loss but without the adversarial loss. The two variants are referred to as HiGAN$\_$Adversarial and HiGAN$\_$CORAL, respectively.  Removing the adversarial loss substantially degrades results, as does removing the CORAL loss. We therefore conclude that both terms are critical to our results. 

\begin{table}
	
	\setlength{\tabcolsep}{5.5mm}{
		\begin{tabular}{|c|c|c|}
			\hline
			Method & S$\rightarrow$U &E$\rightarrow$H  \\
			\hline
			HiGAN$\_$CORAL & 0.726 & 0.335 \\
			HiGAN$\_$Adversarial & 0.921& 0.439  \\
			\hline
			HiGAN & 0.954& 0.446  \\
			
			\hline
	\end{tabular}}
	\caption{Ablation study: classification performance of
		S$\rightarrow$U and E$\rightarrow$H for different losses}
	\label{tab:table4}       
\end{table}

\begin{figure}[htbp]                                                        
	\subfigure[S$\rightarrow$U]{                    
		\begin{minipage}{4cm}\centering                                                          
			\includegraphics[width=1.05\linewidth]{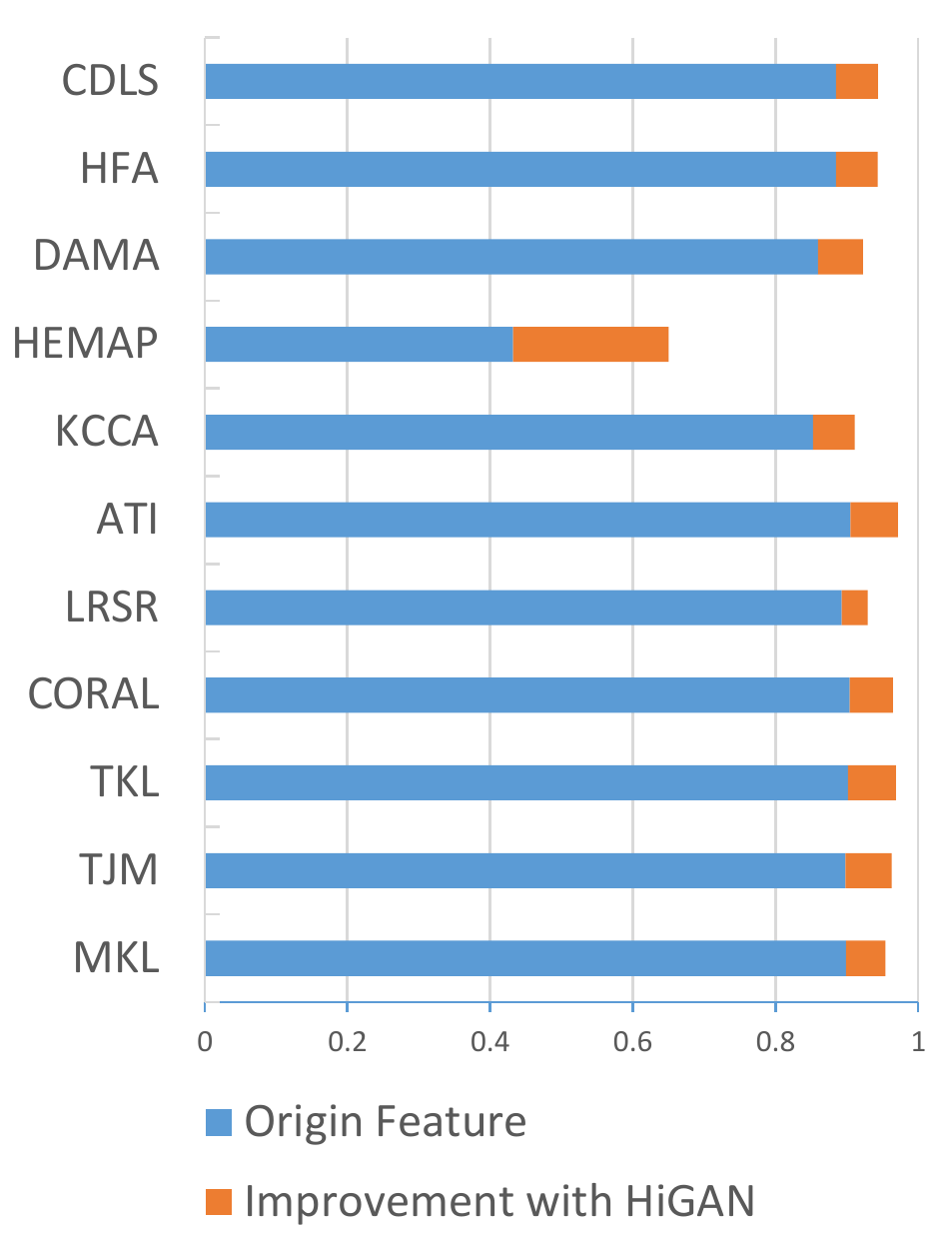}               
	\end{minipage}}\subfigure[E$\rightarrow$H]{                    
		\begin{minipage}{4cm}\centering                                                          
			\includegraphics[width=1.05\linewidth]{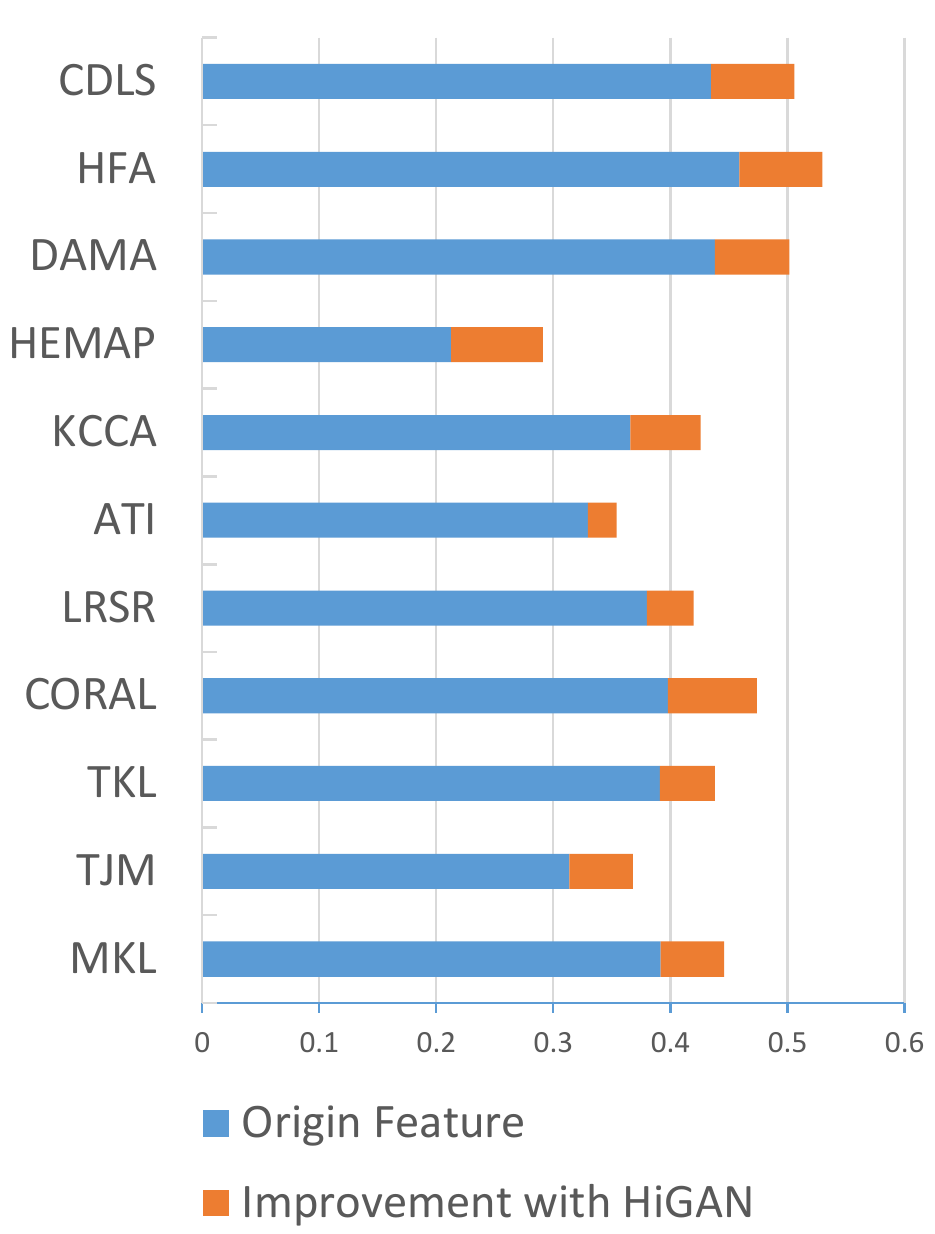}                
	\end{minipage}}\caption{The accuracy improvement in the traditional shallow domain adaptation methods with our HiGAN features on both the S$\rightarrow$U and E$\rightarrow$H datasets. The blue bars show the accuracy of traditional methods trained on JAN and C3D features. The red bars show the absolute increase in accuracy of the traditional methods trained using our HiGAN features.} 
	\label{fig:improvement}                                                        
\end{figure}

\paragraph{Effectiveness of the HiGAN features for traditional domain adaptation.}
\label{para:para1}
To validate the performance improvement of the proposed HiGAN features for video recognition, we compare it with the original features (i.e., JAN and C3D) used in the traditional shallow domain adaptation methods. Additional experiments are conducted with different features. Figure \ref{fig:improvement} (a) and (b) demonstrate the accuracy improvements on the S$\rightarrow$U and E$\rightarrow$H datasets, respectively. We can observe that, the recognition result could be improved with our proposed HiGAN features for both datasets.
\begin{table}
	
	\setlength{\tabcolsep}{4mm}{
		\begin{tabular}{|c|c|c|}
			\hline
			Method & UCF101 &HMDB51  \\
			\hline
			C3D \cite{tran2015learning} & 0.940 & 0.657 \\
			
			HiGAN & 0.980& 0.740  \\
			
			\hline
	\end{tabular}}
	\caption{Comparison between our method and C3D on the video recognition.}
	\label{tab:table3}       
\end{table}
\paragraph{Effectiveness of exploring images by HiGAN for video recognition.}

We also explore whether it is beneficial to transfer knowledge from images for video recognition by comparing our HiGAN features and the C3D features on the UCF101 and the HMDB51 datasets with the chosen categories. The experiments are separately done on the three train/test splits, and the result is averaged across three test splits. From table \ref{tab:table3}, it is interesting to notice that with the same training samples, our method can significantly improve the video recognition accuracy on both datasets, which validates the benefits of exploring related images for video recognition.

\section{Conclusion}

We have proposed a new \textit{Hierarchical Generative Adversarial Networks} approach to transfer knowledge from images to videos. By taking advantage of these two-level adversarial learning with the CORAL loss, our method is capable of learning a domain-invariant feature representation between source images and target videos. Thus the image classifiers trained on the domain-invariant features can be effectively adapted to videos. We conduct experiments on two datasets and the results validate the effectiveness of our method.

The future work includes exploiting large-scale Web images for video recognition which will further improve the recognition accuracy. We also plan to extend our method to multiple source domain adaptation.

\section*{Acknowledgments}
This work was supported in part by the Natural Science Foundation of China (NSFC) under grant No. 61673062 and No. 61472038.

\bibliographystyle{named}
\bibliography{ijcai18}

\end{document}